\documentclass[10pt,twocolumn,letterpaper]{article}

\usepackage{cvpr}
\usepackage{times}
\usepackage{epsfig}
\usepackage{graphicx}
\usepackage{amsmath}
\usepackage{amssymb}
\usepackage[nolist]{acronym}
\usepackage{multirow}
\usepackage{subcaption}
\usepackage{pifont}% http://ctan.org/pkg/pifont
\newcommand{\xmark}{\ding{53}}%

\usepackage{xcolor}

\newcommand{\io}[1]{{\bf #1}}

\usepackage[breaklinks=true,bookmarks=false]{hyperref}

\cvprfinalcopy % *** Uncomment this line for the final submission

\def\cvprPaperID{****} % *** Enter the CVPR Paper ID here

\begin{document}
\newacro{lstm}[LSTM]{Long Short-Term Memory}
\newacro{lsta}[LSTA]{Long Short-Term Attention}
\newacro{clstm}[ConvLSTM]{Convolutional Long Short-Term Memory}
\newacro{rnn}[RNN]{Recurrent Neural Network}
\newacro{cnn}[CNN]{Convolutional Neural Network}
\newacro{tsn}[TSN]{Temporal Segment Network}
\newacro{trn}[TRN]{Temporal Relation Network}
\newacro{tdn}[TDN]{Temporal Difference Network}
\newacro{mfn}[MFNet]{Motion Feature Network}
\newacro{fc}[FC]{Fully Connected}
\newacro{bn}[BN]{Batch Normalization}
\newacro{gru}[GRU]{Gated Recurrent Unit}
\newacro{sgd}[SGD]{Stochastic Gradient Descent}
\newacro{tsm}[TSM]{Temporal Shift Module}
\newacro{flop}[FLOP]{Floating Point Operation}
\newacro{lrcn}[LRCN]{Long-term Recurrent Convolutional Network}
\newacro{gsm}[GSM]{Gate-Shift Module}
\newacro{gsn}[GSN]{Gate-Shift Network}
\newacro{gst}[GST]{Grouped Spatial-temporal Aggregation}

%%%%%%%%% TITLE
\title{FBK-HUPBA Submission to the EPIC-Kitchens Action Recognition\\ 2020 Challenge}

\author{Swathikiran Sudhakaran$^{1}$, Sergio Escalera$^{2,3}$, Oswald Lanz$^{1}$\\[.3cm] %\vspace{-.3cm}
	$^{1}$Fondazione Bruno Kessler - FBK, Trento, Italy\\
	$^{2}$Computer Vision Center, Barcelona, Spain\\
	$^{3}$Universitat de Barcelona, Barcelona, Spain\\
	{\tt\small \{sudhakaran,lanz\}@fbk.eu, \tt\small sergio@maia.ub.es}
}

\maketitle
%\thispagestyle{empty}

%%%%%%%%% ABSTRACT
\begin{abstract}
   In this report we describe the technical details of our submission to the EPIC-Kitchens Action Recognition 2020 Challenge. To participate in the challenge we deployed spatio-temporal feature extraction and aggregation models we have developed recently: \ac{gsm}~\cite{gsm} and EgoACO, an extension of \ac{lsta}~\cite{lsta}. We design an ensemble of GSM and EgoACO model families with different backbones and pre-training to generate the prediction scores. Our submission, visible on the public leaderboard with team name FBK-HUPBA, achieved a top-1 action recognition accuracy of $40.0\%$ on S1 setting, and $25.71\%$ on S2 setting, using only RGB.
\end{abstract}

%%%%%%%%% BODY TEXT
\section{Introduction}
\label{sec:introduction}

% \oswald{'steal' from pami (this here is important) and rephrase} 

Egocentric action recognition is a challenging problem with some differences to third-person action recognition. In egocentric videos, only the hands and the objects that are manipulated under an action are visible. In addition, the motion in the video is a mixture of scene motion and ego-motion caused by the frequent body movements of the camera wearer. This ego-motion may or may not be representative of the action performed by the observer. 
% As a result, existing approaches for third-person action recognition is less suitable in the context of egocentric videos\oswald{not sure this sentence should be in this report}.
Another peculiarity of egocentric videos is that there is typically one `active object' among many in the scene. To identify which among the candidate objects in a cluttered scene is going to be the `active object' requires strong spatio-temporal reasoning. Egocentric action recognition is generally posited as a multi-task learning problem to predict verb, noun and action labels which are related to each other.
% \oswald{add sentence about what is to predict(verb-noun-action)}

To participate in the challenge, we put in place two spatio-temporal feature encoding techniques we have developed, that exhibit some complementarity from a video representation learning perspective:

\begin{itemize}
    \item \ac{gsm}~\cite{gsm} performs early (and deep) spatio-temporal aggregation of features;
    \item EgoACO~\cite{egoACO} performs late (and shallow) temporal aggregation of frame-level or snippet-level features.
\end{itemize}

We have developed variants of both approaches for the challenge, by changing their backbone CNNs and pre-training techniques. We compiled an ensemble out of this pool of trained models to generate the verb-noun-action scores on the test set. Our submission, visible on the public leaderboard at \url{https://competitions.codalab.org/competitions/20115#results}, was obtained by averaging classification scores from ensemble members.

%\swathi{comparison of two models}
\section{Models}
In this section we briefly overview the two model classes utilized for participating in the challenge. More details can be found in the cited papers. Reference implementations of the models in PyTorch framework can be accessed at \url{https://github.com/swathikirans}.
% \oswald{add 1-2 intro sentences}

\subsection{\acl{gsm}}
\label{sec:GSM}
\begin{figure}[h]
	\centering
\includegraphics[width=.82\columnwidth]{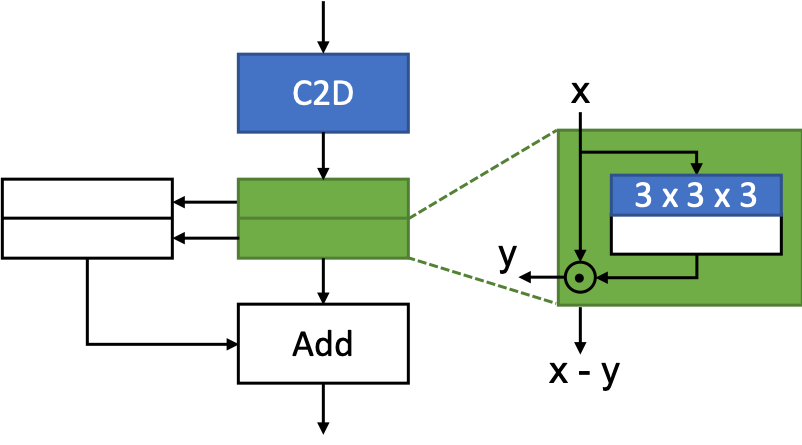}\begin{picture}(0,0)
\put(-194,57){\small \texttt{shift\_fw}}
\put(-194,48){\small \texttt{shift\_bw}}
\put(-35.,47.5){$tanh$}
\vspace{-0.4cm}
\end{picture}
	\caption{Overview of \ac{gsm} (from~\cite{gsm}). 
	}
	\label{fig:gsm}
\end{figure}
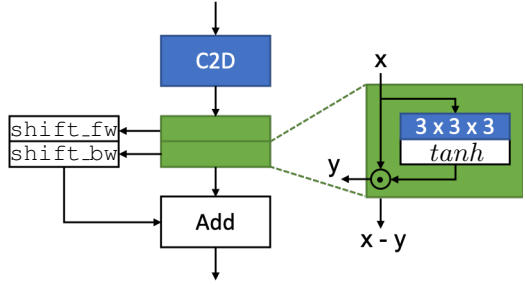

\acf{gsm}~\cite{gsm} is a light weight feature encoding module capable of converting a 2D CNN into an efficient and effective spatio-temporal feature extractor, see Figure~\ref{fig:gsm}.
GSM first applies spatial convolution on the layer input; this is the operation inherited from the 2D CNN base model where GSM is build in.  Then, grouped spatial gating is applied, that is, gating planes are obtained for each of two channel groups, and applied on them.  This separates the 2D convolution output into group-gated features and residual. The gated features are group-shifted forward and backward in time, and zero-padded. These  are finally fused (added) with the residual and propagated to the next layer. This way, GSM selectively mixes spatial and temporal information through a learnable spatial gating. %The implementation of \ac{gsm} is available \hyperlink{https://github.com/swathikirans/GSM}{here}.

% The core of GSM is a gating mechanism that can dynamically route features in a data dependent manner. An overview of GSM is shown in Fig.~\ref{fig:gsm}. In \ac{gsm}, the gating mechanism is implemented via a 3D convolution layer that outputs a single feature plane. The output of the 3D convolution layer is applied to a tanh non-linearity to generate the gate values. The gated features are then temporally shifted in forward and backward directions followed by addition operation to the respective residuals to inject temporal information.

% \oswald{this is too squeezed - again pick a para from the cvpr (copy-paste is fine)} 
A Gate-Shift Network is instantiated by plugging GSM modules inside each of the layers of a 2D CNN, \eg into an Inception style backbone pretrained on ImageNet. The network can be trained for egocentric action recognition in a multi-task setting for verb-noun-action prediction. In the classification layers, we linearly map the action scores into data dependent biases for verb and noun classifiers, as detailed in~\cite{lsta}. %Final predictions are then performed by average pooling the frame-level (now spatio-temporal) scores\oswald{are you sure about this last sentence?}.\swathi{this is from cvpr} %In \ac{gsn}, the network spatio-temporally aggregates the features inside each layer of the backbone CNN.

\subsection{EgoACO}
\label{sec:egoACO}

%\swathi{extension of lsta cvpr, explanation of lsta, attention pooling applied spatially and spatio-temporally, github link}

EgoACO is an extension of our previous conference publication, \ac{lsta}~\cite{lsta}. \ac{lsta} is a recurrent neural unit that extends ConvLSTM with built-in spatial attention and an enhanced output gating. LSTA can be used as frame-level or snippet-level feature sequence aggregator to obtain a video clip descriptor for egocentric action classification. The recurrent spatial attention mechanism enables the network to identify the relevant regions in the input frame or snippet and to maintain a history of the relevant feature regions seen in the past frames. This enables the network to have a smoother tracking of attentive regions. The enhanced output gating constraints LSTA to expose a distilled view of the internal memory. %\oswald{pls review this sentence, pick from cvpr}\swathi{this sentence is from cvpr} enables the network to propagate a filtered version of the memory which is localized on the most discriminative components. 
This allows for a smooth and focused tracking of the latent memory state across the sequence, which is used for verb-noun-action classification. %The implementation of \ac{lsta} is available \hyperlink{https://github.com/swathikirans/LSTA}{here}.

% The  recurrent attention and a novel output pooling mechanism. The attention mechanism allows the model to localize and track the relevant spatial regions in the video while the output pooling facilitates in localizing and propagating the active memory components.\oswald{expand significantly on LSTA description pls}. 

\begin{figure}[h]
	\centering
	\includegraphics[width=\columnwidth]{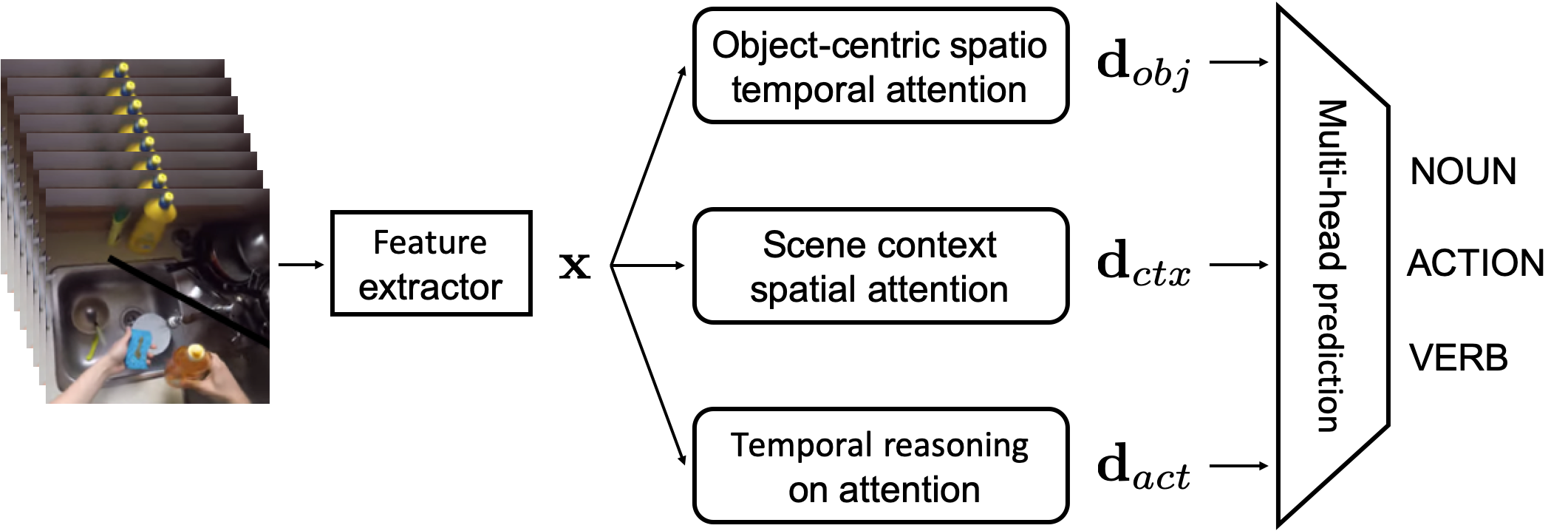}
	\caption{Overview of EgoACO (from~\cite{egoACO}).}
	\label{fig:overview}
\end{figure}

An overview of EgoACO expanding on LSTA is shown in Figure~\ref{fig:overview}.

\noindent EgoACO uses LSTA to perform temporal reasoning to encode a feature sequence $\io{x}$ into an action descriptor $\io{d}_{act}$. Furthermore, the same feature $\io{x}$ is used to generate two other clip descriptors, a scene context descriptor $\io{d}_{ctx}$ and an active object descriptor $\io{d}_{obj}$. The object descriptor encoding applies a frame level or snippet level attention as in~\cite{egornn} but in a temporally coherent manner, and we replace recurrent aggregation with parameter free temporal average pooling. The scene context descriptor $\io{d}_{ctx}$ is generated by applying spatial attention on frame-level features independently, with an average pooled temporal aggregation.  We use $\io{d}_{obj}$ for noun prediction and $\io{d}_{act}$ for verb prediction while all three descriptors are concatenated for action prediction. We apply action score as bias to verb and noun classifiers as done with GSM models.

\begin{figure*}[t]
	\centering
	\includegraphics[width=\linewidth, trim={0 1.5cm 0 0}, clip]{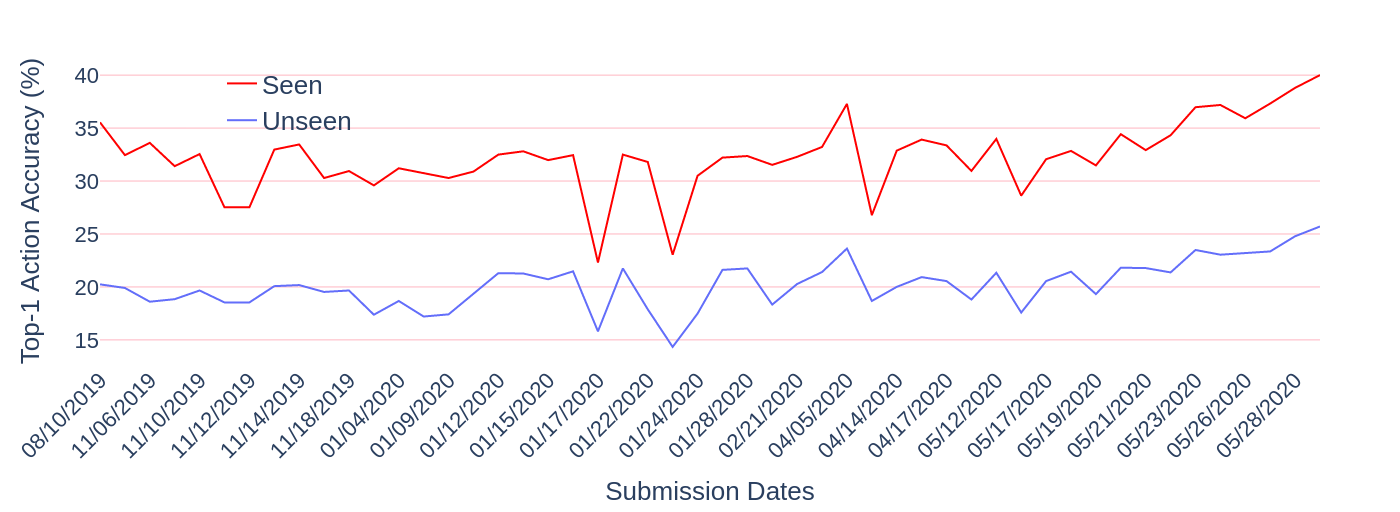}
	\caption{Submission history of our team to the EPIC-Kitchens evaluation server.}
	\label{fig:submission_history}
\end{figure*}

\section{Ensembling Design}
\label{sec:training_details}
%\swathi{comparison of features of the two models}

The clip representations developed by the two model families, \ac{gsm} and EgoACO, are complementary by design. \ac{gsm} performs deep spatio-temporal fusion of features at each layer of a backbone CNN. Thus, the features encoded by \ac{gsm} are highly discriminative for spatio-temporal reasoning. On the other hand, EgoACO relies on the features generated from a pre-trained backbone CNN with narrow temporal receptive field, and learns to combine these frame-level or snippet-level features to develop effective clip level representations. To participate in the challenge, we ensemble models from the two approaches to benefit from their complementarity. We instantiated several variants of the two model families by changing backbone CNNs and using different pre-training strategies. 

\subsection{\bf GSM Variants} 

GSM variants are obtained in the following ways:
\begin{itemize}
	\item Backbone: we used InceptionV3 and BNInception 2D CNNs;
	\item Pre-training: we used ImageNet and EPIC-Kitchens pre-trained models.
\end{itemize}

\noindent{\bf Training.} The entire network is trained end-to-end using SGD with a momentum of 0.9 and a weight decay of $5e^{-4}$. We use a cosine learning rate schedule with linear warmup for 10 epochs. The initial learning rate is 0.01 and the network is trained for 60 epochs. A dropout of 0.5 is used to avoid overfitting. We use random flipping, scaling and cropping for data augmentation. For the EPIC-Kitchens~\cite{Damen2020Collection} pre-trained model, we first trained InceptionV3 with EPIC-Kitchens in a TSN~\cite{tsn} fashion. For EPIC-Kitchens pre-trained BNInception, we used the backbone CNN from~\cite{tbn}. We sample 16 RGB frames from the input video for training and inference. We used the full training set for learning, without validation. We use the model parameters at epoch 60 for testing.

\noindent{\bf Testing.} We perform spatially fully-convolutional inference, following~\cite{non_local}. The shorter side of the frames are scaled to 256 for BNInception and 261 for InceptionV3 backbones. We also sample 2 clips from each video and the predictions of the two clips are averaged to obtain the final score.

\subsection{\bf EgoACO Variants}

EgoACO variants are obtained in the following ways:
\begin{itemize}
	\item Backbone: we used ResNet-34, ResNet-101, ResNet-152, InceptionV3, R(2+1)D ResNet-34, GSM-BNInception and GSM-InceptionV3;
	\item Pre-training: We used ImageNet and IG-65M+Kinetics~\cite{ghadiyaram2019large} pre-trained models.
\end{itemize}

\noindent{\bf Training.} We cloned the top layer of the pretrained CNN three times. The features from these three heads are used for predicting $\io{d}_{obj}$, $\io{d}_{ctx}$ and $\io{d}_{act}$.  We train the network in three stages. In the first stage, the weights of the backbone CNN are frozen while parameters of all the other layers are updated. In stage 2 the parameters of the three heads of the backbone CNN are trained together with the layers trained in stage 1. In stage 3, we train the parameters of \verb|conv4_x| for ResNet based models and the last three layers of Inception based models, together with the layers trained in stage 2. SGD with momentum $0.9$ and weight decay $5e^{-4}$ is used as the optimizer. We use a cosine learning rate schedule for adjusting the learning rate during training. Stages 1 and 2 are trained with an initial learning rate of $0.01$ for 60 epochs while stage 3 is trained for 30 epochs with an initial learning rate of $1e^{-4}$. For \ac{lsta} we use a memory size of 512 and 300 output pooling classes. A dropout of $0.5$ is applied to prevent overfitting. Random scaling, cropping and flipping are applied as data augmentation. $20$ RGB frames sampled from the videos are applied as input to the network for the 2D CNN based backbones. For R(2+1)D based model, we sample $20$ snippets consisting of $3$ RGB frames while for GSM based models, we sample $16$ RGB frames as input. We used the full training set for learning, without validation. We use the model parameters at epoch 30 of stage 3 for testing.

\noindent{\bf Testing.} We use the center crop of the video frame, along with 2 clip sampling strategy during inference.

\section{Results and Discussion}
\label{sec:results}

%Using the same feature representation as input to the three feature descriptors, represented in Fig.~\ref{fig:overview} is suboptimal. Using three separate CNN is computationally expensive and can lead to overfitting due to the massive parameter overhead. In our approach, we share the layers of the backbone except the top most layer and separate the top layer for the three different heads. 

%The baseline model consists of a 2D CNN pre-trained with ImageNet as the feature extractor. We developed different variants of EgoACO by changing the feature extractors. We used ResNet-34, ResNet-101, ResNet-152 and InceptionV3 2D CNNs. In addition to the 2D CNN backbones, we also used a R(2+1)D ResNet-34 backbone. In this model, we used a temporal striding of 1 in layers \verb|conv4_x| and \verb|conv5_x| of the backbone. The input to the models are a sequence of frame stacks consisting of 3 frames. The R(2+1)D backbone is pre-trained on IG-65M and Kinetics datasets. We also used GSM-BNInception and GSM-InceptionV3 backbones as feature extractors.

\begin{table*}[t]
	\centering
	\begin{tabular}{c|c|c|c|c|c}
		Method & Backbone & Pre-training & Ensemble 1 & Ensemble 2 & Ensemble 3 \\ \hline \hline
		\multirow{5}{*}{EgoACO} & IncV3 & ImageNet & \checkmark & \checkmark & \xmark \\ \cline{2-6}
		& Res-34 & ImageNet & \checkmark & \checkmark & \xmark \\ \cline{2-6}
		& Res-101 & ImageNet & \checkmark & \checkmark & \checkmark \\ \cline{2-6}
		& Res-152 & ImageNet & \xmark & \xmark & \checkmark \\ \cline{2-6}
		& R(2+1)D Res-34 & IG-65M + Kin. & \xmark & \checkmark & \checkmark \\ \hline \hline
		\multirow{3}{*}{GSM} & BNInc & EPIC & \checkmark & \checkmark & \checkmark \\ \cline{2-6}
		& IncV3 & ImageNet & \checkmark & \checkmark & \xmark \\ \cline{2-6}
		& IncV3 & EPIC & \xmark & \checkmark & \checkmark \\ \hline \hline
		\multirow{2}{*}{GSM+EgoACO} & BNInc & EPIC & \checkmark & \checkmark & \xmark \\ \cline{2-6}
		& IncV3 & EPIC & \xmark & \checkmark & \checkmark \\ \hline \hline
		Ensemble 1 & - & - & - & \xmark & \checkmark \\ \hline
		Ensemble 2 & - & - & \xmark & - & \checkmark \\ \hline
	\end{tabular}
	\caption{List of individual models used as part of the ensembles.}
	\label{tab:ensemble}
\end{table*}

Figure~\ref{fig:submission_history} tracks our submission history to the EPIC-Kitchens evaluation server. We started off this year's challenge with our third place score from EPIC-Kitchens Action Recognition 2019 challenge. Our participation to the challenge can be divided into three phases. The first one corresponds to the development of \ac{gsm} followed by the development of EgoACO. The third phase concerns with the development of the different variants of the two model families and their ensembling. Our final score is from the last submission.

Table~\ref{tab:ensemble} lists the models that were used as part of our ensemble. We chose the models based on the variability in pre-training, backbones and feature encoding approaches. The best single model is EgoACO with R(2+1)D ResNet-34 backbone that achieves a top-1 action accuracy of 37.32\% on S1 and 23.35\% on S2 setting. The result obtained from the submission server for each of the three ensembles is listed in Table~\ref{tab:epic_kitchens}. Model ensembling is done by averaging the prediction scores of the individual models. Our best performing ensemble, ensemble 3, results in a top-1 action accuracy of \textbf{40.0\%} and \textbf{25.71\%} in seen and  unseen settings, respectively.

%In this technical report, we described the details of our approach for participating in the action recognition task of the EPIC-Kitchens 2020 challenge. We used two different families of models that differ in the feature encoding approach. We developed variants of the two model families using different backbones and pre-training strategies. An ensemble of the models from the two families result in a top-1 action accuracy of 40.0\% and 25.71\% in Seen and Unseen settings, respectively.

%\swathi{mention fig 2}

\begin{table*}[th]\small
	\centering
	\begin{tabular}{c|l|c|c|c|c|c|c|c|c|c|c|c|c}
		\hline
		& \multirow{2}{*}{Method} & \multicolumn{3}{c|}{Top-1 Accuracy (\%)} & \multicolumn{3}{c|}{Top-5 Accuracy (\%)} & \multicolumn{3}{c|}{Precision (\%)} & \multicolumn{3}{c}{Recall (\%)} \\
		\cline{3-14}
		& & \parbox{0.7cm}{Verb} & \parbox{0.7cm}{Noun} & \parbox{0.8cm}{Action} & \parbox{0.7cm}{Verb} & \parbox{0.7cm}{Noun} & \parbox{0.8cm}{Action} & \parbox{0.7cm}{Verb} & \parbox{0.7cm}{Noun} & \parbox{0.8cm}{Action} & \parbox{0.7cm}{Verb} & \parbox{0.7cm}{Noun} & \parbox{0.8cm}{Action}\\
		\hline \hline
		\multirow{3}{*}{\rotatebox[origin=t]{90}{S1}}  & Ensemble 1  & 66.98 & 47.21 & 37.28 & 89.71 & 71.05 & 58.18 & 57.91 & 43.25 & 18.84 & 45.41 & 43.96 & 22.70 \\ \cline{2-14}
		
		& Ensemble 2  & 68.41 & 48.76 & 38.80 & 90.54 & 72.41 & \textbf{60.41} & 59.64 & 44.74 & 21.40 & \textbf{47.45} & 45.56 & 23.78 \\ \cline{2-14}
		
		& \textbf{Ensemble 3}  & \textbf{68.68} & \textbf{49.35} & \textbf{40.00} & \textbf{90.97} & \textbf{72.45} & 60.23 & \textbf{60.63} & \textbf{45.45} & \textbf{21.82} & 47.19 & \textbf{45.84} & \textbf{24.34} \\ \cline{2-14}
	 \hline
        \multirow{3}{*}{\rotatebox[origin=t]{90}{S2}}
		& Ensemble 1  & 54.15 & 31.38 & 23.63 & 80.81 & 56.54 & 41.99 & 27.72 & 24.43 & 12.02 & 23.50 & 28.46 & 17.71\\ \cline{2-14}
		
		& Ensemble 2  & 55.96 & 32.23 & 24.79 & 81.56 & 58.83 & 44.14 & \textbf{32.97} & 26.43 & \textbf{12.84} & \textbf{25.16} & \textbf{29.56} & \textbf{18.31}\\ \cline{2-14}
		
		& \textbf{Ensemble 3}  & \textbf{56.67} & \textbf{33.90} & \textbf{25.71} & \textbf{81.87} & \textbf{59.68} & \textbf{44.42} & 30.72 & \textbf{27.25} & 12.74 & 25.09 & 29.46 & 17.93\\ 
		\hline 
	\end{tabular}
	\caption{Comparison of recognition accuracies obtained with various ensembles of models in EPIC-Kitchens dataset.}
	\label{tab:epic_kitchens}
\end{table*}

\section*{Acknowledgements} This research was supported by AWS Machine Learning Research Awards and ICREA under the ICREA Academia programme.

{\small
\bibliographystyle{unsrt}
\bibliography{epic20_bib}

\begin{thebibliography}{1}

\bibitem{gsm}
Swathikiran Sudhakaran, Sergio Escalera, and Oswald Lanz.
\newblock {Gate-Shift Networks for Video Action Recognition}.
\newblock In {\em Proc. CVPR}, 2020.

\bibitem{lsta}
Swathikiran Sudhakaran, Sergio Escalera, and Oswald Lanz.
\newblock {LSTA: Long Short-Term Attention for Egocentric Action Recognition}.
\newblock In {\em Proc. CVPR}, 2019.

\bibitem{egoACO}
Swathikiran Sudhakaran, Sergio Escalera, and Oswald Lanz.
\newblock {Learning to Reason about Actions on Objects in Egocentric Video with
  Attention Dictionaries}.
\newblock {\em arXiv 2020}.

\bibitem{egornn}
Swathikiran Sudhakaran and Oswald Lanz.
\newblock {Attention is All We Need: Nailing Down Object-centric Attention for
  Egocentric Activity Recognition}.
\newblock In {\em Proc. BMVC}, 2018.

\bibitem{Damen2020Collection}
Dima Damen, Hazel Doughty, Giovanni~Maria Farinella, Sanja Fidler, Antonino
  Furnari, Evangelos Kazakos, Davide Moltisanti, Jonathan Munro, Toby Perrett,
  Will Price, and Michael Wray.
\newblock {The {EPIC-KITCHENS} Dataset: Collection, Challenges and Baselines}.
\newblock {\em IEEE Transactions on Pattern Analysis and Machine Intelligence
  (TPAMI)}, 2020.

\bibitem{tsn}
Limin Wang, Yuanjun Xiong, Zhe Wang, Yu~Qiao, Dahua Lin, Xiaoou Tang, and Luc
  Van~Gool.
\newblock {Temporal Segment Networks for Action Recognition in Videos}.
\newblock {\em IEEE Transactions on Pattern Analysis and Machine Intelligence
  (TPAMI)}, 41(11):2740--2755, 2018.

\bibitem{tbn}
Evangelos Kazakos, Arsha Nagrani, Andrew Zisserman, and Dima Damen.
\newblock {Epic-Fusion: Audio-Visual Temporal Binding for Egocentric Action
  Recognition}.
\newblock In {\em Proc. ICCV}, 2019.

\bibitem{non_local}
Xiaolong Wang, Ross Girshick, Abhinav Gupta, and Kaiming He.
\newblock {Non-local Neural Networks}.
\newblock In {\em Proc. CVPR}, 2018.

\bibitem{ghadiyaram2019large}
Deepti Ghadiyaram, Du~Tran, and Dhruv Mahajan.
\newblock Large-scale weakly-supervised pre-training for video action
  recognition.
\newblock In {\em Proc. CVPR}, 2019.

\end{thebibliography}
}

\end{document}